# Recalling of Images using Hopfield Neural Network Model

Ramya C[1], Kavitha G[2] and Dr. Shreedhara K S[3]
[1]M.Tech (Final Year), [2]Lecturer and [3]Professor, Dept. of Studies in CS&E
University B.D.T College of Engineering, Davangere
Davangere University, Karnataka, INDIA
[1]cramyac@gmail.com

*Abstract*— In the present paper, an effort has been made for storing and recalling images with Hopfield Neural Network Model of auto-associative memory. Images are stored by calculating a corresponding weight matrix. Thereafter, starting from an arbitrary configuration, the memory will settle on exactly that stored image, which is nearest to the starting configuration in terms of Hamming distance. Thus given an incomplete or corrupted version of a stored image, the network is able to recall the corresponding original image. The storing of the objects has been performed according to the Hopfield algorithm explained below. Once the net has completely learnt this set of input patterns, a set of testing patterns containing degraded images will be given to the net. Then the Hopfield net will tend to recall the closest matching pattern for the given degraded image. The simulated results show that Hopfield model is the best for storing and recalling images.

## I. INTRODUCTION

An Artificial Neural Network (ANN) is an information processing paradigm that is inspired by the biological nervous systems, such as the brain. It is composed of a large number of highly interconnected processing elements (neurons) working in unison to solve specific problems. A Neural Network is configured for pattern recognition or data classification, through a learning process. In biological systems, Learning involves adjustments to the synaptic connections that exist between the neurons. Neural networks process information in a similar way the human brain does. The network is composed of a large number of highly interconnected processing elements working in parallel to solve a specific problem. Neural networks learn by example. A neuron has many inputs and one output. The neuron has two modes of operation (i) the training mode and (ii) the using mode. In the training mode, the neuron can be trained for particular input patterns. In the using mode, when a taught input pattern is detected at the input, its associated output becomes the current output. If the input pattern does not belong in the taught list of input patterns, the training rule is used.

The earliest recurrent neural network has independently begun with Anderson (1977), Kohonen (1977), but Hopfield (1982) presented a complete mathematical analysis of such a subject [4]. That is why this network is generally referred to as the Hopfield network. It is a fully connected neural network model of associative memory in which we can store information by distributing it among neurons, and recall it from the neuron states dynamically relaxed.

The conventional Hopfield model is the most commonly used model for auto-association and optimization. Hopfield networks are auto-associators in which node values are iteratively updated based on local computation principle: the new state of each node depends only on its net weighted input at a given time. This network is fully connected network and the weight matrix determination is one of the important tasks while using it for any application.

In the following sections, a complete Hopfield Neural Network model has been presented along with its structure and main characteristics. The algorithm for storing and recalling images has been given. Experimental results show the stored set of images, degraded images for testing the net and corresponding original images for the given corrupted images as output. Hence, we say Hopfield net is the best for storing and recalling images.

## II. LITERATURE SURVEY

With the introduction in 1982 of the model named after him, John Hopfield established the connection between neural networks and physical systems of the type considered in statistical mechanics. This in turn gave computer scientists a whole new arsenal of mathematical tools for the analysis of neural networks. Other researchers had already considered more general associative memory models in the 1970s, but by restricting the architecture of the network to a symmetric connection matrix with a zero diagonal, it was possible to design recurrent networks with stable states. With the introduction of the concept of the energy function, the convergence properties of the networks could be more easily analyzed.

The Hopfield network also has the advantage, in comparison with other models, of a simple technical implementation using electronic or optical devices. The computing strategy used when updating the network states corresponds to the relaxation methods traditionally used in physics. The properties of Hopfield networks have been investigated since 1982 using the theoretical tools of statistical mechanics. Gardner published a classical treatise on the capacity of the perceptron and its relation to the Hopfield model. The total field sensed by particles with a spin can be computed using the methods of mean field theory. This simplifies a computation which is hopeless without the help of some statistical assumptions. Using these methods Amit et al. showed that the number of stable states in a Hopfield network of *n* units is bounded by 0.14*n*. A *recall* error is tolerated only 5% of the time. This upper bound is one of the most cited numbers in the theory of Hopfield networks.

In 1988 Kosko proposed the BAM model, which is a kind of "missing link" between conventional associative memories and Hopfield networks. Many other variations have been proposed since, some of them with asynchronous, others with synchronous dynamics. Hopfield networks have also been studied from the point of view of dynamical systems. In this respect spin glass models play a relevant role. The efforts of Hopfield and Tank with the TSP led to many other similar experiments in related fields. Wilson and Pawley repeated their experiments but they could not confirm the optimistic results of the former authors. The main difficulty is that complex combinatorial problems produce an exponential number of local minima of the energy function.

The Hopfield model and its stochastic variants have been applied in many other fields, such as psychology, simulation of ensembles of biological neural networks, and chaotic behavior of neural circuits. The optical implementation of Hopfield networks is a promising field of research. Other than masks, holograms can also be used to store the network weights. The main technical problem is still the size reduction of the optical components, which could make them a viable alternative to conventional electronic systems.

### III. PRESENTATION OF THE HOPFIELD NETWORK

The Hopfield network consists of a set of N interconnected neurons which update their activation values asynchronously and independently of other neurons. All neurons are both input and output neurons. The Hopfield net is shown in Fig.1.

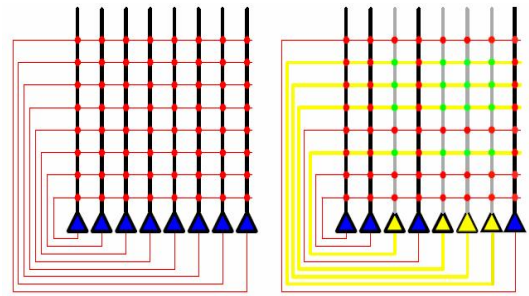

Figure. 1 Hopfield Neural network, Synapses (green) between active neurons (yellow) are strengthened.

The set of input vectors (input data, learning data, training vectors, input patterns) is a set of vectors of *N* the form

$$I(x) = \{x(1); x(2);…; x(N)\}, x(k) \in H^n, k \in 1,N$$
$$x(k) = (x_1(k)x_2(k) … x_n(k))^T$$

where *H* is the Hamming space having only the elements -1 and +1 (bipolar elements) and *T* means the transposition. Originally, Hopfield chose for each *x(k)* the binary activation values 1 and 0, but using bipolar values +1 and -1 presents some advantages [1,4]. For an input vector *x* we denote by $x^c \in H^n$ the complement, where the value 1 from *x* becomes -1 in $x^c$ and vice versa. So, if we encode *I(x)*, we also encode its complement $I(x^c)$. The Hamming distance between two vectors having the same type $u,v \in H^n$ is a function $DH[u,v]$ representing the number of bits that are different between *u* and *v*. The Hamming distance *DH* is related to the Euclidian distance *DE* by the equation $DE = 2\sqrt{(DH)}$ [2].

*Main Characteristics of Hopfield Network:*

1. The first aim of the Hopfield network is **to store** the input data *I(x)* (store phase).
2. Then, the second aim is **to recall** one input vector from *I(x)* if one knows a test vector *xr*, with $xr \in H^n$ (retrieval phase). Generally, a test vector is a noisy vector of *I(x)*. This characteristic makes the Hopfield network useful for restoring degraded images [5]. Hence the Hopfield network is used to solve a store-recall problem.

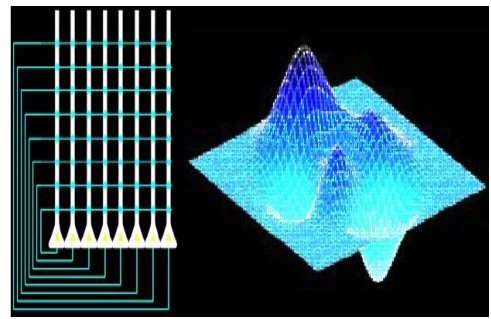

Figure. 2 How memories are recalled

3. In the Hopfield network there are weights associated with the connections between all the neurons of layer $S_x$. We organize the weights in a square and symmetric matrix
$$W = W_{nxn}, \quad W = (w_{ij}), \quad w_{ij} = w_{ji}.$$
All connections inside the layer $S_x$ are bidirectional and the units may, or may not, have feedback connections to themselves. We can take $w_{ii} = 0$. We denote the columns of $W$ by vectors $w_j \in R^n$, $j = 1, n$ and the lines of $W$ by vectors $line_i \in R^n$, $i = 1, n$.

4. The Hopfield network is a fully interconnected network.

5. The matrix $W$ can be determined in advance if all the training vectors $I(x)$ are known.
We use the formula
$$W = \sum_{k=1}^{N} x(k)x(k)^T, \quad w_{ii} = 0$$
Formula (1)

Here it is possible to know the input vectors $x(k)$ one after one. Here it is necessary to know from the beginning all the input vectors $x(k)$. This matrix does not change in time. After the matrix $W$ was computed, we say the input data $I(x)$ was stored.

6. Having the matrix $W$ and a test vector $xr$ we look for the corresponding training vector. In order to find the corresponding vector of $I(x)$ we use the Hopfield Algorithm (see below).

7. we call the energy function associated with the input vector $x \in H^n$ and the matrix $W$ the real number
$$E = E(x), \quad E(x) = -x^T W x$$
$$E(x) = -\sum_{i=1}^{n}\sum_{j=1}^{n} x_i w_{ij} x_j$$

The energy function is called also Lyapunow function in the theory of dynamical systems.

IV. THE HOPFIELD ALGORITHM

From the known input set $I(x)$, the Hopfield Algorithm retrieves a input vector with the help of a test vector $xr$ (noisy vector). At the beginning we denote the test vector by $xr = xr(1)$. At the time $t$ (natural number), the algorithm propagates the information inside the $S_x$ layer as follows:
a) compute activation
$$net_j(t) = <w_j, xr(t)>, \quad j = \overline{1,n}$$
Formula (2)
where $w_j$ is the column $j$ of the matrix W.
b) update the values on $S_x$ layer, namely the vector $xr(t)$ becomes $xr(t+1)$ with the components [2,4].

$$xr_j(t+1) = \begin{cases} +1, & net_j(t) > 0 \\ -1, & net_j(t) < 0 \\ xr_j(t), & net_j(t) = 0 \end{cases}, \quad j = \overline{1,n}.$$
Formula (3)

The Hopfield has the following steps.

*Step 1.* One knows the input set $I(x)$ and store it by computing the matrix $W = W_{nxn}$ with formula (1). One or more test vectors $xr$ are given. Put $t = 1$.

*Step 2.* (optional) Compute the complement set $I(x^c)$ and the Hamming distance between the vectors of $I(x)$ and the test vectors $xr$. Check if the input set $I(x)$ is orthogonal or not.

*Step 3.* At the time $t$, propagate the information from the $S_x$ layer to the $S_x$ layer by the formulas (2) and (3). So we find the vector $xr(t+1) \in H^n$.

*Step 4.* Put $t + 1$ instead of $t$ and repeat the step 3 until there are no further changes in the components of $xr(t + 1)$.

*Note:* If all goes well, the final stable state will recall one of the vectors $I(x)$ used to construct the store matrix $W$. The final output of the Hopfield Algorithm is a vector which is closest in Hamming distance to the original vector $xr$.

V. EXPERIMENTAL RESULTS

We have conducted experiments to illustrate the storing and recalling of bitmap images using Hopfield NN model. We used 3-D arrays to store both input patterns and testing patterns. We present simulator output in Fig. 3. In which a) shows the stored set of bitmap images which the network learns. Once network has learnt successfully, we give some incomplete or corrupted bitmap images which are shown in b). Hopfield net recalls the images which are stored in it and gives the corresponding original image which will be the closest matching pattern for the given incomplete or corrupted image which is shown in c).

VI. CONCLUSIONS

The present study has demonstrated that a Hopfield network is the best model for storing and recalling images. Images are stored by calculating a corresponding weight matrix. Hopfield net recalls the images which are stored in it and gives the corresponding original image which will be the closest matching pattern for the given incomplete or corrupted image.

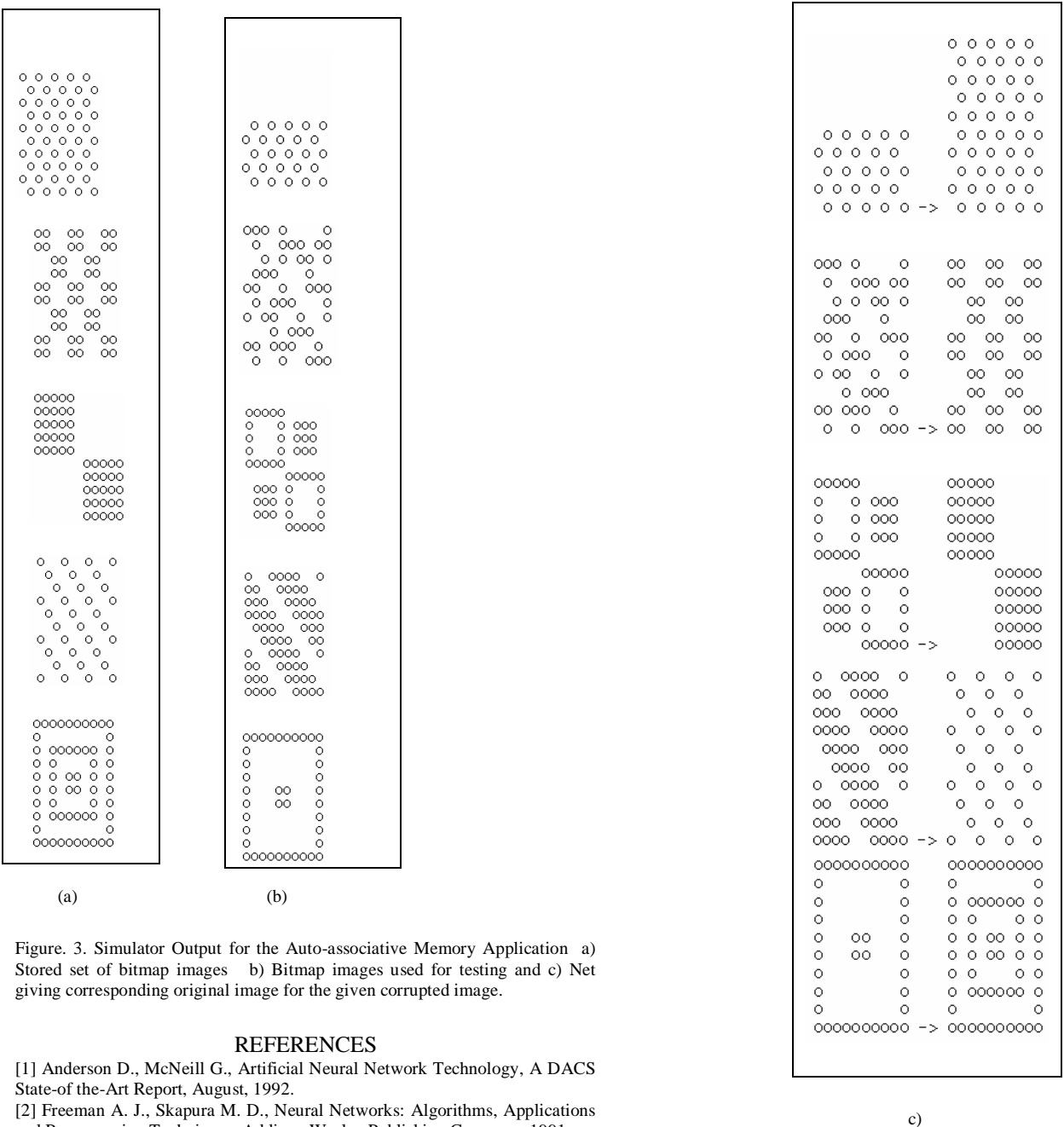

(a)          (b)

Figure. 3. Simulator Output for the Auto-associative Memory Application  a) Stored set of bitmap images   b) Bitmap images used for testing and c) Net giving corresponding original image for the given corrupted image.

c)